\documentclass[a4paper,sort&compress, fleqn]{cas-sc}

\usepackage[numbers]{natbib}

\usepackage{textcomp}

\usepackage{subfigure}
\usepackage{url}
\usepackage{verbatim}
\usepackage{graphicx}
\usepackage{amssymb}
\usepackage{amsmath}
\usepackage{array,booktabs}
\usepackage{multirow}
\usepackage{bm}
\usepackage{bbding}
\usepackage{pifont}
\usepackage{makecell}
\usepackage{xcolor}
\usepackage{array}
\usepackage{tabularx}
\usepackage{newfloat}
\usepackage{listings}
\usepackage{algorithm}
\usepackage{algorithmic}
\usepackage{hyperref}
\usepackage{subcaption}
\usepackage{multicol}

\def\tsc#1{\csdef{#1}{\textsc{\lowercase{#1}}\xspace}}
\tsc{WGM}
\tsc{QE}
\tsc{EP}
\tsc{PMS}
\tsc{BEC}
\tsc{DE}

\begin{document}
\let\WriteBookmarks\relax
\def\floatpagepagefraction{1}
\def\textpagefraction{.001}
\let\printorcid\relax
\shorttitle{}

\shortauthors{Liao et~al.}

\title [mode = title]{Real-time Accident Anticipation for Autonomous Driving Through Monocular Depth-Enhanced 3D Modeling}                      

\author[1]{Haicheng Liao}
\fnmark[1]
\credit{Conceptualization, Methodology, Experiment, Writing}

\author[2]{Yongkang Li}
\credit{Methodology,Experiment}
\fnmark[1]

\author[3]{Chengyue Wang}
\ead{zhenningli@um.edu.mo}
\credit{Methodology}

\author[4]{Songning Lai}
\ead{zhenningli@um.edu.mo}
\credit{Methodology}

\author[5]{Zhenning Li}
\cormark[1]
\ead{zhenningli@um.edu.mo}
\credit{Conceptualization, Methodology, Writing}

\author[6]{Zilin Bian}
\credit{Methodology}

\author[7]{Jaeyoung Lee}
\credit{Writing}

\author[8]{Zhiyong Cui}
\credit{Methodology}

\author[9]{Guohui Zhang}
\credit{Methodology}

\author[1]{Chengzhong Xu}
\credit{Conceptualization, Funding, Review}

\affiliation[1]{organization={State Key Laboratory of Internet of Things for Smart City and Department of Computer and Information Science, University of Macau}, city={Macau SAR}, country={China}}

\affiliation[2]{organization={Department of Information and Software Engineering, University of Electronic Science and Technology of China}, city={Chengdu}, country={China}}

\affiliation[3]{organization={State Key Laboratory of Internet of Things for Smart City, University of Macau}, city={Macau SAR}, country={China}}

\affiliation[4]{organization={State Key Laboratory of Internet of Things for Smart City and Departments of Civil and Environmental Engineering and Computer and Information Science, University of Macau}, city={Macau SAR}, country={China}}

\affiliation[5]{organization={Thrust of Artificial Intelligence, Hong Kong University of Science and Technology (Guangzhou)}, city={Guangzhou}, country={China}}

\affiliation[6]{organization={ Transportation Planning and Engineering in the Department of Civil and Urban Engineering, New York University}, city={New York}, country={United States}}

\affiliation[7]{organization={School of Traffic and Transportation Engineering, Central South University}, city={Changsha}, country={China}}

\affiliation[8]{organization={School of Transportation Science and Engineering, Beihang University}, city={Beijing}, country={China}}

\affiliation[9]{organization={Department of Civil and Environmental Engineering, University of Hawaii}, city={Honolulu HI}, country={United States}}

\cortext[cor1]{Corresponding author; $^{1}$Equally Contributed}

\begin{abstract}
The primary goal of traffic accident anticipation is to foresee potential accidents in real time using dashcam videos, a task that is pivotal for enhancing the safety and reliability of autonomous driving technologies. In this study, we introduce an innovative framework, AccNet, which significantly advances the prediction capabilities beyond the current state-of-the-art 2D-based methods by incorporating monocular depth cues for sophisticated 3D scene modeling. Addressing the prevalent challenge of skewed data distribution in traffic accident datasets, we propose the Binary Adaptive Loss for Early Anticipation (BA-LEA). This novel loss function, together with a multi-task learning strategy, shifts the focus of the predictive model towards the critical moments preceding an accident.  {We rigorously evaluate the performance of our framework on three benchmark datasets—Dashcam Accident Dataset (DAD), Car Crash Dataset (CCD), and AnAn Accident Detection (A3D), and DADA-2000 Dataset—demonstrating its superior predictive accuracy through key metrics such as Average Precision (AP) and mean Time-To-Accident (mTTA).} 

\end{abstract}


\maketitle
\begin{sloppypar}

\section{Introduction}
\label{sec:intro}
The surge in global traffic accidents has escalated traffic safety into a paramount public health issue \cite{li2024understanding}. The World Health Organization (WHO) highlights an alarming statistic: annually, 1.45 million lives are claimed by traffic accidents, with injuries surpassing 50 million \cite{mannering2016unobserved}. By 2030, traffic accidents are projected to become the fifth leading cause of death globally \cite{li2024efficient}. The financial impact of these accidents is tremendous, costing more than \$180 billion annually in the United States alone. This economic burden, coupled with the loss of life, underscores the urgency of enhancing traffic safety measures \cite{khan2024advancing}. This has become one of the most important and challenging tasks for traffic management authorities \cite{ahmed2023accounting,li2023mitigating,li2024steering}. 

In response to this challenge, accident anticipation, the predictive analysis of vehicular collisions before they occur is a pivotal development for enhancing the safety protocols of autonomous driving \cite{fang2023vision,ahmed2020analysis}. This field has garnered significant interest due to its potential to elevate the safety measures of intelligent vehicle systems dramatically \cite{guan2024world,li2024context,tang2024adaptive}. Effective anticipation, even seconds before a potential accident, can empower autonomous systems to initiate critical safety maneuvers, potentially averting collisions \cite{ali2024advances,kolekar2020human,liao2024mftraj}.

\begin{figure}
    \centering
    \includegraphics[width=0.8\linewidth]{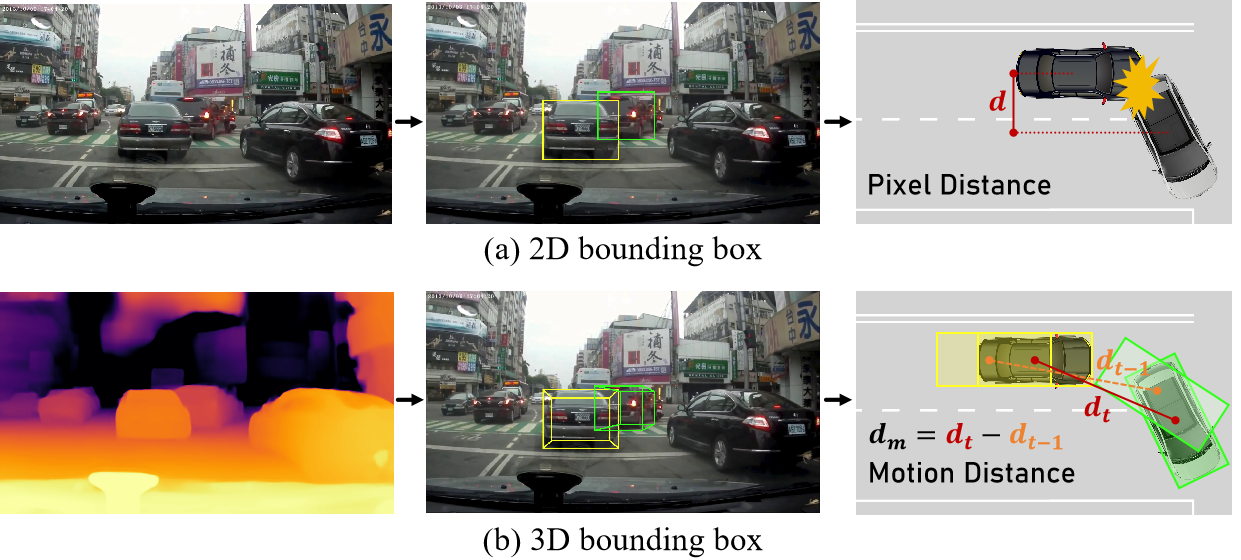}
    \caption{ {Illustration of Monocular Depth-Enhanced 3D Modeling in AccNet. Current methods (a) all rely on 2D bounding boxes identified by object detectors and typically resort to pixel-by-pixel distance computations to approximate motion interactions in driving scenarios, leading to inaccuracies due to the lack of depth information. In contrast, our AccNet model (b) extracts precise 3D coordinates of key traffic participants, including vehicles and pedestrians, from video and leverages monocular depth to calculate real-world distances. This method enhances accident detection by capturing detailed fine-grained correlations between visual images over time.}}
    \label{head}
\end{figure}

Yet, mastering accident anticipation poses substantial challenges, primarily due to the complexity and the often limited and noisy visual data in an observed dashcam video \cite{Zeng_2017_CVPR}. Traffic scenes, viewed from the perspective of the onsite camera, are cluttered with various elements, such as cars, pedestrians, and motorcyclists. Within this busy visual landscape, cues critical to predicting accidents can be lost amidst irrelevant data, making it difficult for systems to detect impending collisions, especially in complex intersections. Despite these challenges, advancements in modeling, particularly those focusing on uncertainty, hold promise in distinguishing between pertinent and extraneous visual information, identifying potential risks based on variables like irregular vehicle movements \cite{chen2022review,yu2024online}.

 {
As shown in Fig. \ref{head}, traditional methods \cite{yao2022dota,fang2022traffic,BaoICCV2021DRIVE} in accident anticipation have leaned heavily on analyzing two-dimensional pixel distances within video frames, utilizing these as proxies for the actual spatial distances between objects. This approach, however, neglects the crucial depth information, vital for a comprehensive understanding of the spatial interactions and movements that define vehicular and pedestrian dynamics. For example, in Fig. \ref{head} (a), using a 2D bounding box may lead to a misjudgment of a collision due to the visual alignment of vehicles, whereas in Fig. \ref{head} (b) using a 3D bounding box allows for the correct assessment that the vehicles have not collided. This example underscores a critical inquiry: \textit{Do current methodologies adequately incorporate the three-dimensional spatial dynamics essential for precise accident prediction?}}

 {
Furthermore, the issue of data imbalance in traffic datasets poses a significant challenge for deep learning models, which rely on balanced data distributions for effective pattern recognition. Since videos containing accidents often include numerous non-accident clips, this imbalance favors non-accident scenarios, substantially diminishing the models' exposure to, and recognition of, the rare yet crucial instances preceding an accident. Additionally, the introduction of depth information may enhance the existing data bias. These factors lead to another critical question: \textit{How can we refine predictive models to highlight the ephemeral, yet pivotal moments that precede an accident?} The limited availability of accident data further impacts the models' ability to generalize, increasing the risk of overlooking actual accidents.}

In response to these challenges, we present a novel framework designed to enhance the anticipation of traffic accidents in complex traffic environments. Central to our approach is the development of a 3D Collision Module \cite{Mertan_2022,Yin2023Metric3DTZ}, which leverages monocular depth information to derive precise three-dimensional coordinates of critical entities like vehicles and pedestrians. This module ensures a more accurate depiction of spatial relationships and dynamics captured in dashcam footage, fostering a deeper understanding of potential collision scenarios. Concurrently, we address the issue of imbalanced data distribution by introducing the Binary Adaptive Loss for Early Anticipation (BA-LEA). This innovative loss function refocuses the model on the decisive moments leading up to an accident, augmented by a multi-task learning approach that fine-tunes the balance between different loss functions to optimize overall model performance. Additionally, the implementation of a Smooth Module promotes training stability, smoothing out fluctuations and bolstering the model's capacity to assimilate complex data sets effectively.

To sum up, our contributions are threefold:

1) We developed a module that leverages monocular depth information to extract precise 3D coordinates of objects from images, enhancing spatial dynamics understanding in dashcam videos. This module also introduces a novel graph topology for Graph Neural Networks (GNN), improving risk assessment.

2) Our innovative loss function, tailored for complex traffic situations, emphasizes important pre-accident indicators. It's supported by a multi-task learning strategy and a Smooth Module, which together ensure optimized, stable learning.

3)  {Through rigorous testing on DAD, CCD, A3D, and DADA-2000 datasets, our model demonstrates superior performance in key metrics like Average Precision (AP) and mean Time-To-Accident (mTTA), outperforming existing approaches and marking a significant advancement in accident anticipation technology.}

\section{Related Work} 

The problem of accident anticipation in autonomous driving systems presents a formidable challenge, requiring the detection of collisions from dashboard camera footage and the accurate prediction of their occurrence. Introduced by Chan et al. \cite{chan2016anticipating} in 2016, this domain extends beyond traditional accident prediction by necessitating a deep understanding of the complex and unpredictable nature of traffic accidents. The rarity of crash events, coupled with the dynamic and variable nature of traffic scenes, significantly complicates this task. 

In response to these challenges, a spectrum of studies \cite{liu2023net,rahim2021deep,huang2020highway,hussain2022hybrid,zhang2022real} have employed Convolutional Neural Networks (CNNs) in tandem with sequential networks like Recurrent Neural Networks (RNNs), Long Short-Term Memory (LSTM) units, and Gated Recurrent Units (GRUs). For example, Yao et al. \cite{yao2019unsupervised} combine CNN and RNN to predict the temporal features of scenes leading up to accidents, thereby identifying clues in videos that point to future incidents. Takimoto et al. \cite{takimoto2019predicting} apply GRU and RNN networks to predict accidents by considering deviations in the future positions of traffic agents. Furthermore, Basso et al. \cite{basso2021deep} propose a new image-inspired architecture that consists of CNNs to capture the microscopic scene of vehicle behavior. Thakare et al. \cite{thakare2023rareanom} propose a convolutional autoencoder designed for detection and classification, providing an efficient means of extracting key features while minimizing computational demands. Yet, this approach encounters limitations in identifying long-distance patterns, especially in scenarios involving sparse objects. This combination facilitates the processing of time-series visual data and the extraction of complex motion features, representing a significant stride in capturing the temporal dynamics of traffic scenes. Further advancements in this field have seen the incorporation of attention mechanisms  \cite{Karim2021stdan,vaswani2017attention,BaoICCV2021DRIVE,karim2023attention}, which refine the model's ability to distill complex temporal interactions.  Furthermore,  in light of the burgeoning success of Transformers in computer vision, numerous researchers have explored transformer-based models in this field. These models leverage self-attention mechanisms to capture interrelationships within time-series trajectory data, which minimizes information loss and enables the network to learn and extract long-range dependencies.   {For instance, UniFormerv2 \cite{li2022uniformerv2}, VideoSwin \cite{liu2022video}, CAVG \cite{liao2024gpt}, and MVITv2  \cite{fan2021multiscale} all employ the transformer-based model to delineate the relative positioning of traffic agents such as vehicles and pedestrians, underscoring the potential of Transformers in enhancing model understanding of dynamic traffic scenes.}

Despite their proficiency in handling temporal data, these frameworks often fail to accurately capture spatial relationships, a critical aspect when interpreting visually dense traffic scenes. To bridge this gap, researchers have turned to Graph Convolutional Networks (GCNs) \cite{BaoMM2020,zhao2019t,Thakur_2024_WACV,liao2024crash}, leveraging the spatial positioning of traffic agents to construct graph-based representations. Notable work such as Song et al. \cite{song2024dynamic} present a dynamic attention-augmented graph network for accident anticipation and Wang et al. \cite{wang2023gsc} introduce a spatio-temporal graph convolutional network in this domain. These approaches, utilizing bounding boxes to outline agents, facilitate a more nuanced understanding of spatio-temporal dynamics, exemplified by the uncertainty-based GCN proposed by Bao et al. \cite{BaoMM2020} and the innovative GCRNN framework by Wang et al. \cite{wang2023gsc}, which crafts 2D coordinates to represent missing agents in the visual field. Notably, Liao et al. \cite{liao2024and} enhance accident anticipation by incorporating accident localization through the use of LLMs for in-depth scene analysis, allowing for accurate warnings regarding the nature, timing, and location of possible incidents.

Despite these advancements, a predominant reliance on GCRNN-based frameworks, which process video frames sequentially, has been observed. This methodological approach not only impedes the efficiency of model training and inference but also approximates the distance between objects using two-dimensional pixel distances, neglecting the critical depth aspect. Such limitations underscore a pressing need for models that can accurately capture the three-dimensional spatial dynamics of traffic scenes.

Inspired by the burgeoning field of monocular depth estimation \cite{Mertan_2022,Roberts_2021_ICCV,Yin2023Metric3DTZ}, our work introduces a novel model, AccNet, built upon the foundation of ZoeDepth \cite{bhat2023zoedepth}. This model innovatively combines monocular depth information with pixel distances, transcending the conventional 2D analysis limitations. By accurately extracting the absolute 3D coordinates of objects such as vehicles, pedestrians, and buildings from images, AccNet offers a precise representation of spatial dynamics in traffic accident scenes. This capability not only enhances the reconstruction of 3D crash scenarios but also marks a significant leap forward in the domain of crash anticipation, setting a new benchmark for future research.

\section{Methodology}
\subsection{Problem Formulation}
The advancement of automated systems in identifying potential traffic accidents through dashcam footage is pivotal in enhancing road safety. Our model, leveraging dashcam videos, aims to predict traffic accidents by analyzing sequential frames, thereby providing early warnings that could be crucial in preventing misshapes.
  
Given a video sequence $V = \{V^0, V^1, \ldots, V^T\}$ consisting of $T$ frames, our task is to assign a probability score $s^t$ to each frame, indicating the likelihood of an accident occurring at or after that frame. If an accident is anticipated to occur at frame $\tau$, we define the Time-to-Accident (TTA) as $\Delta t = \tau - t^\theta$, where $t^\theta$ is the earliest frame at which the probability score $s^t$ exceeds a predefined threshold $s^\theta$. Thus, a video is classified as containing an accident (positive) if $s^t \geq s^\theta$ for any $t \geq t^\theta$. Otherwise, it is considered accident-free (negative), setting $\tau=0$. Our goal is to enhance the accuracy of predicting such incidents and maximize the TTA, providing more lead time for preventive actions.

\begin{figure}[t]
    \centering
    \includegraphics[width=0.9\linewidth]{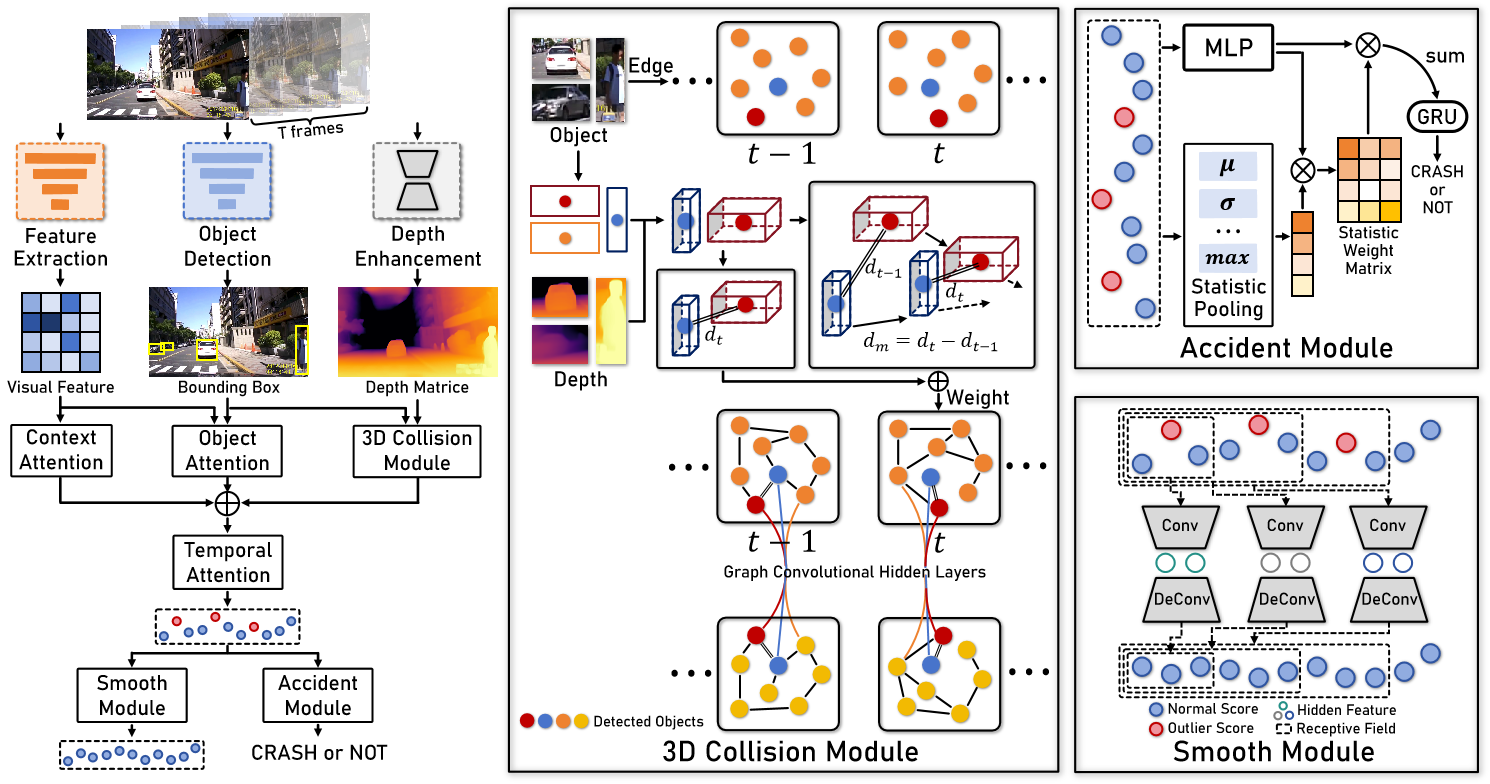}
    \caption{ {Overall architecture of the proposed AccNet. The feature extractor, object detector, and depth enhancer first generate the visual features, bounding boxes, and depth matrices respectively for the raw T-frame video sequence. These outputs are then fed into the proposed Context Attention, Object Attention, and 3D Collision modules to update the representation of object queries. The Temporal Attention mechanism then fuses the features from the three modalities. Finally, the Smooth Collision Module is introduced to balance multi-task learning, while the Accident Module predicts the likelihood of an accident occurring at each time step.}}
    \label{structure}
\end{figure}

\subsection{Model Architecture}
As illustrated in Fig. \ref{structure}, our proposed model architecture, termed AccNet, integrates feature extraction, attention mechanisms, and a novel 3D collision detection module to analyze dashcam video data effectively. At its core, AccNet aims to discern intricate patterns that precede accidents, enabling early prediction and extending the TTA window.
\subsubsection{Feature Extraction and Depth Enhancement}
 {
The initial phase involves processing the video frames through a series of pre-trained models for feature extraction. Utilizing MMDetection \citep{mmdetection}, we identify traffic agents (vehicles, pedestrians) in each frame $V^t$, focusing on the top $N$ objects based on confidence scores. We then crop images $\{V^t_1, V^t_2, ..., V^t_N\}$ from the bounding boxes of these identified objects, employing VGG-16 \cite{Simonyan15} to extract comprehensive context $H$ from $V^t$ and visual features $O=\{O_{1}, O_{2}, \cdots, O_{N}\}$ from $\{V^t_1, V^t_2, ..., V^t_N\}$. To capture the spatial depth, we leverage the ZoeDepth model \cite{bhat2023zoedepth}, an advanced tool for depth estimation, to construct depth matrices $D^t$ for each frame $t$. This multi-faceted approach ensures a rich feature set, ready for subsequent analysis. }

\subsubsection{Context Attention and Object Attention}
Context Attention and Object Attention are designed to prioritize context $H$ and object $O$ features from video frames. The Context Attention employs a multi-head attention framework, translating the context features into query $Q_I$, key $K_I$, and value $V_I$ components through Multilayer Perceptrons (MLP). These components are then segmented into $h$ distinct heads $Q_I^g$, $K_I^g$, and $V_I^g$, and $g \in [1, h]$, producing the context vector $I_C$. Mathematically,
\begin{equation}
I_C = \sum_{g=1}^h \text{head}^g = \sum_{g=1}^h \phi_{\text{softmax}}\left(\frac{Q_I^g \cdot K_I^g}{\sqrt{d^g}}\right) \cdot V_I^g
\end{equation}

Here, $\phi_{\text{softmax}}$ denotes the softmax activation function. $d^g = \frac{d}{h}$ and $d$ represent the dimension size of $Q_I^g$ divided across the heads. In parallel, the Object Attention dynamically allocates attention across the detected objects using a series of learnable weight matrices $W_\theta$, $W_\beta$, and $W_b$, which yields a weighted representation $W_o$ and then results in the object vector $I_O$, thus enhancing the model's focus on pertinent objects. Formally,
\begin{equation}
I_O = \phi_{sum} [W_o O]=\phi_{sum} [\underbrace{\phi_{\text{softmax}}(W_\beta^T \cdot \phi_{\tanh}(W_\theta O + W_b))}_{W_o} O]
\end{equation}
where $\phi_{\tanh}$ is the tanh activation function, and $\phi_{sum}$ represents the sum of the vectors of the detected objects.

\subsubsection{3D Collision Module}
Existing work \cite{wang2023gsc,Thakur_2024_WACV} typically constructs graph weights using the pixel distance between target objects to learn the latent spatial relationships between agents. However, these approaches often overlook the significant discrepancy between pixel distances and actual distances. To address this issue, we introduce an innovative 3D Collision Module that transcends traditional pixel distance metrics. 

Specifically, the pre-extracted depth information $D^t$ is fed to construct a GCN under 3D settings. For the $n$-th object in the $t$-th frame, the position expression $p^t_{n}$ of the object detection bounding box can be represented as follows:
\begin{equation}
p^t_{n}=[p^t_{1,n}, p^t_{2,n}]=[\underbrace{(x^t_{1,n},y^t_{1,n})}_{p^t_{1,n}}, \underbrace{(x^t_{2,n},y^t_{2,n})}_{p^t_{2,n}}]
\end{equation}
where $p^t_{1,n}$ and $p^t_{2,n}$ are the pixel coordinates of the top-left and bottom-right coordinates of the bounding box, respectively. Furthermore, given the center coordinates of the bounding box $c^t_{x,n} = \frac{({x^t_{1,n}}+{x^t_{2,n}})}{2}$ and $c^t_{y,n} = \frac{({y^t_{1,n}}+{y^t_{2,n}})}{2}$, the depth at the target object's center point can be applied as its depth information $D^t(c^t_{x,n},c^t_{y,n})$. 

Hence, the constructed three-dimensional coordinates of the target can be represented as $P^t_{n}=(c^t_{x,n},c^t_{y,n},D^t(c^t_{x,n},c^t_{y,n}))$. 
For target objects ${i,j}\in n$, we define the distance between target objects at time $t$ as: $d^t_{\{i,j\}}=||P^t_{i}-P^t_{j}||^2$.
In real-world driving scenarios, collisions are often related to the approach and retreat movements of vehicles. To assess this, it is necessary to measure the velocity of two vehicles in adjacent frames. In this study, for target object $n$, the velocity between two consecutive frames is represented as follows:
\begin{equation}
\overrightarrow{P^t_{n}}=(c^{t}_{x,n}-c^{t-1}_{x,n},c^t_{y,n}-c^{t-1}_{y,n},D^t(c^t_{x,n},c^t_{y,n})-D^{t-1}(c^{t-1}_{x,n},c^{t-1}_{y,n}))
\end{equation}
Consequently, the motion difference between two targets $i,j$ can be calculated as follows:
\begin{equation}
m^t_{i,j}=\frac{\overrightarrow{P^t_{i}}}{||\overrightarrow{P^t_{i}}||}-\frac{\overrightarrow{P^t_{j}}}{||\overrightarrow{P^t_{j}}||}
\end{equation}
Alternatively, the edge weight is defined as $q^t_{\{i,j\}}=\lambda_1d^t_{\{i,j\}}+\lambda_2m^t_{\{i,j\}}$, where $\lambda_1+\lambda_2=1$. This formulation of graph weights incorporates both the relative positional information and the motion trends between targets. To balance the magnitude of values between different targets,  the edge weight $q^t_{\{i,j\}}$ is further normalized as follows:
\begin{equation}
W_{\{i,j\}} = \frac{e^{q^t_{\{i,j\}}}}{\sum_{i,j\in n}e^{q^t_{\{i,j\}}}}
\end{equation}
Next, we introduce a GCN with a novel topology, which is constructed as follows:
\begin{equation}\label{eq.7-1}
{G} = \{V,{E}\}=\phi_{\text{GCN}}(C \oplus O, E, W)
\end{equation}
where $G$ is the spatial vector output from this module, $\phi_{\text{GCN}}$ represents the GCN, and $\oplus$ denotes concatenation. $W=\{W_{\{i,j\}}, \forall i, j \in [1,N], \text{ and } i \neq j \}$ is the normalized matrix. Correspondingly, \( {E}\) is the set of edges illustrating potential connections between objects. Objects are considered connected if they are identified by the detector. Additionally, the node set's futures comprise the context $V$ and object $C$ vectors. 

Overall, this module computes the real-world distance of an object by extending the depth information to extract fine-grained correlations between multimodal input vectors, focusing on the object's inherent motion properties. 

\subsubsection{Temporal Attention}
The Temporal Attention is primarily responsible for extracting the temporal features from the context $C$, object $O$, and spatial $G$ vectors. The objective of this task is to calculate the likelihood of an accident happening in each frame of a video sequence. To accomplish this, we combine these vectors and apply a hyperbolic tangent function. This is followed by multiplication with a weight matrix $W_m$ and subsequent normalization with a softmax activation function, culminating in the generation of an attention matrix $W_a$ over the feature dimensions. The attention matrix $W_a$ plays a critical role in balancing the interrelationships between the context $C$, object $O$, and spatial $G$ vectors, generating a focused feature fusion $W_f$. This process is expressed by the following equations:
\begin{equation}
\begin{aligned}
W_a=softmax(tanh(I_s \oplus O_s \oplus G) \circ W_m), \quad W_f=(I_s \oplus O_s \oplus G)\circ W_a
\end{aligned}
\end{equation}
where $\circ$ is the element-wise product operator. Following the attention mechanism, we feed $W_f$ into a Gated Recurrent Unit (GRU) for temporal processing. To mitigate overfitting during training, we incorporate a two-layer ``dropout-linear'' configuration, with dropout rates set to 0.5 and 0.1, respectively. 

\subsubsection{Smooth Module}
It is critical to distinguish between genuinely hazardous situations and those that, while potentially dangerous, do not result in accidents. Notably, certain maneuvers by vehicles or pedestrians, although aggressive or risky, may not lead to accidents but still cause significant spikes in the model's predicted probability scores $s^t$, crossing predefined thresholds $s^\theta$ and erroneously flagging false positives. 

As previously stated, it is important for this task to differentiate between hazardous situations and those that are potentially dangerous but do not result in accidents. Certain maneuvers by vehicles or pedestrians, although aggressive or risky, may not lead to accidents but can still cause significant spikes in the model's predicted probability scores $s^t$. This can result in false positives if the predefined thresholds $s^\theta$ are crossed. These instances, occurring within a brief time window ($\delta = t_2 - t_1$), highlight the need for a refined analytical approach. To mitigate this, we introduce a sophisticated Smooth Module, as shown in Fig. \ref{structure}, employing convolutional and deconvolutional operations across variable temporal scales. This technique allows for the downsampling of video data to unearth latent features $F_{\delta}$ within these critical intervals, subsequently upsampled to their original temporal resolution $F_{\delta}^0$. The application of this method over scales of 2, 5, and 10 seconds achieves comprehensive smoothing of feature volatility, effectively distinguishing between transient high-risk behaviors and actual precursors of accidents. The temporally smoothed features are then harmonized with the original input via a residual connection, at a mixing ratio of 1:0.15, further enhancing the model's ability to accurately interpret dynamic road scenarios without overfitting to sporadic high-risk maneuvers.

\subsubsection{Accident Module}
The Smooth Module refines the model's sensitivity to false triggers, while the Accident Module uses insights from the Temporal Attention Module to determine the occurrence of accidents in analyzed video sequences. This module employs a novel statistical attention mechanism, which differs from traditional queries by utilizing statistical pooling layers. These layers are adept at extracting key probabilistic features—mean $\overline{S_h}$, variance $\sigma(S_h)$, maximum value $max(S_h)$, and range $\Delta = S_h - \overline{S_h}$—from the output probability scores of each video. By integrating statistical dimensions using a specialized attention framework, the module synthesizes a statistical weight matrix that reflects the nuanced interplay of predictive scores over time. This matrix informs a GRU and subsequent linear layers, resulting in a refined model capable of discerning the likelihood of accidents with greater precision and reliability.

\subsection{Training}
Our training loss function is divided into two main components: the probability score loss function $\mathcal{L}_S$ and the prediction loss function $\mathcal{L}_p$, which aim to optimize the model's output probability scores $S$ and improve the accuracy of traffic accident detection, respectively. 
 
Traditionally, the loss in probability score is computed using a cross-entropy formula that aligns the output probability scores of positive and negative videos with their expected outcomes. These outcomes are close to 1 for accidents and close to 0 for non-accidents. However, this binary classification does not fully account for temporal variability and the gradual evolution of risk across video frames. To tackle the issue of temporal inconsistency in accident prediction, we present the Binary Adaptive Loss for Early Anticipation (BA-LEA) method, which is an extension of the AdaLEA framework \cite{Suzuki2018AnticipatingTA}. The BA-LEA method adjusts the loss calculation by considering the temporal proximity to potential accidents, thereby enhancing the model's sensitivity to the unfolding dynamics within video sequences. It employs two distinct temporal coefficients, $\lambda_1$ and $\lambda_2$, for positive and negative instances, respectively, which are meticulously calibrated to enhance prediction accuracy by adjusting the penalty based on the temporal distance from the observed incident: For positive instances (indicative of accidents), $\lambda_1=e^{-\max \left( \frac{\tau - t}{f_1}, 0 \right)}$ is designed, where $\tau$ represents the actual occurrence time of an accident, and $f_1$ serves as a decay factor. This coefficient increases the penalty for deviation from the expected probability score as the frame approaches the accident time, ensuring heightened model responsiveness during critical moments preceding an accident. Conversely, for negative instances (where no accident occurs), $\lambda_2=\frac{t}{f_2}$ progressively amplifies the penalty for scores straying towards accident prediction as time advances, addressing the challenge of overestimating the likelihood of accidents in benign scenarios. This dual-coefficient strategy ensures a dynamic recalibration of the model's focus, prioritizing temporal segments closer to accident events in positive videos and enhancing vigilance over time in negative videos to avoid false alarms. This nuanced approach is encapsulated in the revised loss function, which can be defined as follows:
\begin{equation}
\mathcal{L}_S = \frac{1}{V} \sum_{v=1}^{V} \left( -l_v \sum_{t=1}^{T} e^{-\max \left( \frac{\tau - t}{f_1}, 0 \right)} \log(s_{t}) - (1 - l_v) \sum_{t=1}^{T} \frac{t}{f_2}\log(1 - s_{t}) \right)
\end{equation}
where $V$ represents the total number of videos, and $l_v$ denotes the label of the video, with $l_v=1$ for positive videos and $l_v=0$ for negative ones.

Complementing $\mathcal{L}_S$, the prediction loss function $\mathcal{L}_p$ directly influences the model's accuracy in distinguishing between accident and non-accident scenarios. This is achieved by penalizing deviations from the true accident occurrence within the training dataset, encouraging the model to refine its predictive capabilities. Formally,
\begin{equation}
\mathcal{L}_p = \frac{1}{V} \sum_{v=1}^{V} \left[ -l_v \log(l_{p}) - (1 - l_v) \log(1 - l_{p}) \right]
\end{equation}
where $l_p$ is the prediction output from the Accident Module. In traditional approaches, the sum of the loss functions $\mathcal{L}_S$ and $\mathcal{L}_p$ is scaled by a coefficient to adjust their relative importance.

The fusion of $\mathcal{L}_S$ and $\mathcal{L}p$ into a cohesive loss function poses a challenge, given the necessity to balance their contributions effectively. Inspired by multitask learning principles and the work of Kendall et al. \cite{kendall2018multitask}, we adopt a multitask loss modulation strategy. This approach introduces uncertainty coefficients $\sigma_1$ and $\sigma_2$, allowing for the dynamic adjustment of each loss component's influence during the training process. The operations are formalized as:
\begin{equation}
\mathcal{L}=\frac{1}{2\sigma_1^2}\mathcal{L_S}+\frac{\gamma}{2\sigma_2^2}\mathcal{L}_{p}+log(\sigma_1\sigma_2)
\end{equation}
where $\gamma$ is a predetermined coefficients, with $\sigma_1$ and $\sigma_2$ initially set to 1. Both $\sigma_1$ and $\sigma_2$ could dynamically adjust the contribution of each task's loss during training. This loss function construction ensures accurate model predictions on a video level while also enabling early accident prediction on a frame level.

\section{Experiment}
\subsection{Experiment Setup}
\subsubsection{Datasets}
1) \textbf{Dashcam Accident Dataset (DAD)} \cite{chan2016anticipating}. The DAD dataset comprises 620 dashcam videos that were recorded across six major cities in Taiwan. Each video lasts 5 seconds and has a frame rate of 20. From these videos, we extracted 1750 segments, consisting of 620 positive segments (accidents) and 1130 negative segments (non-accidents). Accidents occur at the 90th frame in the positive segments. The dataset is split into a training set (70\%) and a testing set (30\%). The training set comprises 455 positive and 829 negative videos, while the testing set consists of 165 positive and 301 negative videos.

2) \textbf{Car Crash Dataset (CCD)} \cite{BaoMM2020}.  {The CCD dataset is a comprehensive collection of environmental attributes, including day/night and snow/rain/clear weather conditions, as well as information on bicycles and pedestrians involvement, and detailed descriptions of accident causes. The dataset consists of 4500 videos, each 5 seconds long with a frame rate of 10 frames per second. The accident videos in the dataset are derived from full-length videos, and while there is no uniform accident occurrence time across videos, all accidents occur within the last 2 seconds of the video. The dataset is divided into training and testing sets, with 80\% for training and 20\% for testing. Both sets maintain a 1:2 ratio of positive to negative videos.}

3) \textbf{AnAn Accident Detection} \cite{yao2018egocentric}. The A3D dataset comprises 1500 dashcam videos obtained from various cities in East Asia, recorded under different weather conditions and at different times of day. Each video lasts 5 seconds, with 20 frames per second through downsampling. Accidents are marked at the 80th frame for positive segments. The dataset is divided into 80\% for training and 20\% for testing.

 {
4) \textbf{DADA-2000 Dataset} \cite{Fang2019dada}. The DADA-2000 dataset comprises 2,000 accident videos, each averaging approximately 230 frames in length at a rate of 30 frames per second. It is the first dataset dedicated to predicting driver attention in driving accident scenarios. DADA-2000 encompasses diverse weather conditions, lighting, road environments, and types of accident participants. We divided the dataset into training and testing sets at a ratio of 4:1.}

\subsubsection{Evaluation Metrics}
In the context of traffic accident anticipation tasks, three evaluation metrics are paramount: Average Precision (AP), Area Under the Curve (AUC) and mean Time-To-Accident (mTTA).

1) \textbf{AP}: AP measures the model's ability to accurately distinguish between positive (accident) and negative (non-accident) videos, essentially assessing the model's detection capability for traffic accidents within videos. For binary classification tasks, we denote $\textit{TP}$ (true positives), $\textit{FP}$ (false positives), and $\textit{FN}$ (false negatives) to represent the instances of correctly predicted accidents, incorrectly predicted accidents, and missed accidents, respectively. The model's recall is given by $R = \frac{\textit{TP}}{\textit{TP} + \textit{FN}}$, and precision is $P = \frac{\textit{TP}}{\textit{TP} + \textit{FP}}$. $\textit{AP}$ is defined as the area under the precision-recall curve, mathematically expressed as $AP = \int P(R) dR$.  { For the A3D dataset, given its lack of non-accident data, we employ accuracy as a substitute for AP values.}

 {
2) \textbf{AUC}: AUC quantifies the probability that a randomly selected positive instance is ranked higher than a randomly selected negative instance by the predictive model. Unlike AP, which measures the precision across various levels of recall, AUC assesses the model's capability to distinguish between positive and negative classes. It is mathematically defined as the area under the Receiver Operating Characteristic (ROC) curve, represented by the formula: \( AUC = \int R(F) \, dF \), where \( R(F) \) denotes the recall function of the false positive rate \( F \).}

3) \textbf{mTTA}: mTTA measures the model's capability to predict the occurrence of an accident in positive samples ahead of time. If an accident occurs at frame $\tau$, TTA is defined as $\Delta t = \tau - t_\theta$, where $t_\theta$ satisfies $s_t \geq s_\theta$ for $t \geq t_\theta$ and $s_t < s_\theta$ for $t < t_\theta$, with $s_\theta$ being the threshold. For all different thresholds $s_\theta \in [0,1]$, mTTA is defined as the mean of all TTAs, i.e., $\textit{mTTA} = \frac{1}{n} \sum_{s_{\theta}}\textit{TTA}$. 

\subsubsection{Implementation Details}
The study utilized Pytorch to implement the proposed method and conducted training and testing on an A40 48G GPU. Pre-trained models with features extracted by VGG-16 with a dimension of 4096 were used. Depth information from videos was extracted using the ``ZoeD-M12-NK'' version of the ZoeDepth model. Regarding training parameters, the learning rate of the model is set to $1\times10^{-4}$ with a batch size of 16. We use ReduceLROnPlateau as the learning rate scheduler to ensure that each model undergoes at least 50 training epochs. For the loss function parameters, the decay coefficients are set as $f_1=20, f_2=150$, and the proportional coefficient of the loss function $\gamma$ is $1\times10^{-3}$. Further specifics regarding the implementation and essential parameter settings of our model are provided as follows:

\textbf{Feature Extraction and Depth Enhancement.} These two modules receive video $V$ as input, with dimensions of $(B, T, X, Y)$, where $X$ and $Y$ correspond to the width and height of the video in pixels. The feature extraction process generates a feature matrix with dimensions of $(B, T, N_0, D)$, where $B$ represents the batch size, $T$ represents the total number of frames in the video, $N_0$ is the sum of the number of target detection objects $N$ and the overall image feature, thus $N_0=N+1$, and $D$ signifies the feature hidden dimension after feature extraction by VGG-16. The output of the depth enhancement process has the same dimensions $(B, T, X, Y)$ as the original input video. In this implementation, $N$ is set to 19, which is the maximum number of detected objects.

\textbf{Context Attention and Object Attention.} The matrix $F$ undergoes a linear transformation that reduces the hidden dimension $D$ to $D_h$. The input matrix is divided into object vector $H \in (B, T, N, D)$ and context vector $O \in (B, T, 1, D)$. Furthermore, $H$ and $O$ are then separately processed by Context Attention and Object Attention to obtain outputs $H_c$ and $O_o$ with dimensions $(B, T, 1, D_c)$ and $(B, T, 1, D_o)$, respectively. The input matrix is divided into object features $H$ with dimensions $(B, T, N, D)$ and context features $O$ with dimensions $(B, T, 1, D)$. Then, $H$ and $O$ are separately processed by Context Attention and Object Attention to obtain outputs $H_c$ and $O_o$ with dimensions $(B, T, 1, D_c)$ and $(B, T, 1, D_o)$, respectively. Additionally, a multi-head attention mechanism with 8 heads is employed in these components.

\textbf{3D Collision Module.} The input depth features with dimensions $(B, T, X , Y)$ along with context features $O$ are fed into the 3D Collision Module, resulting in an output $F_d$ with dimensions $(B, T, 1, D_g)$. The $\lambda_1$ and $\lambda_2$ are set to 0.6 and 0.4, respectively.

 {
\textbf{Temporal Attention.} The outputs from Context Attention, Object Attention, and the 3D Collision Module are concatenated into a mixed feature \(F_m\) with dimensions \((B, T, 1, H_c + O_o + F_d)\). In our implementation, \(H_c\), \(O_o\), and \(F_d\) are set to 512, 256, and 256, respectively. \(F_m\) is subsequently passed into the Temporal Attention module to produce an output with dimensions \((B, T, 2)\). Within the Temporal Attention, \(F_m\) is first fed into a GRU with a hidden dimension of \(H_T = 512\), yielding a hidden feature tensor of dimensions \((B, T, H_T)\), which is then passed to a linear layer for dimension transformation. This output is then forwarded to both the Smooth Module and the Accident Module for further processing.}

 {
\textbf{Accident Module.} Similar to the Temporal Attention, the Accident Module takes \(F_m\) as input and computes four statistical measures of the feature dimensions, concatenating them into a tensor of dimensions \((B, T, 4)\). This tensor undergoes dimension transformation via linear layers, then is input into a GRU with hidden dimensions \(H_A = 32\), and finally through a linear transformation to produce an output of dimensions \((B, 2)\).}

\subsection{Comparison to State-of-the-art (SOTA) Baselines}
\begin{table}[!ht]
  \centering
  \caption{ {Comparison of models seeking \textbf{balance} between mTTA and AP on four datasets. \textbf{Bold} and \underline{underlined} values represent the best and second-best performance in each category. Instances where values are not available are marked with a dash (``-'').}}
    \setlength{\tabcolsep}{2mm}
    \resizebox{\linewidth}{!}{
    \begin{tabular}{ccccccccc}
    \toprule
    \multirow{2}[2]{*}{Model} & \multicolumn{2}{c}{DAD} & \multicolumn{2}{c}{CCD} & \multicolumn{2}{c}{A3D} & \multicolumn{2}{c}{DADA-2000}\\
    \cmidrule(l{3pt}r{3pt}){2-3} \cmidrule(l{3pt}r{3pt}){4-5} \cmidrule(l{3pt}r{3pt}){6-7} \cmidrule(l{3pt}r{3pt}){8-9} \multicolumn{1}{c}{} & AP(\%)$\uparrow$ & mTTA(s)$\uparrow$ & AP(\%)$\uparrow$ & mTTA(s)$\uparrow$ & AP(\%)$\uparrow$ & mTTA(s)$\uparrow$ & AP(\%)$\uparrow$ & mTTA(s)$\uparrow$\\
    \midrule 
        DSA \cite{DSA2016Chan}                & 48.1 & 1.34 & 98.7 & 3.08 & 92.3 & 2.95 & - & - \\
        ACRA \cite{Zeng2017CVPR}              & 51.4 & 3.01 & 98.9 & 3.32 & - & - & - & - \\
        AdaLEA \cite{Suzuki2018AnticipatingTA}& 52.3 & 3.44 & 99.2 & 3.45 & 92.9 & 3.16 & 54.13 & 3.64 \\
        DSTA \cite{Karim2021stdan}            & 52.9 & 3.21 & 99.1 & 3.54 & 93.5 & 2.87 & 58.24 & 3.32 \\
        UString \cite{BaoMM2020}              & 53.7 & 3.53 & \textbf{99.5} & 3.74 & 93.2 & 3.24 & 61.72 & 3.59 \\
        GSC \cite{wang2023gsc}                & \underline{58.2} & 2.76 & \underline{99.3} & 3.58 & \underline{94.9} & 2.62 & \underline{64.95} & 3.66 \\
        XAI-Accident \cite{karim2022xai} & 52.0 & \textbf{3.79} & 94.0 & \textbf{4.57} & 92.8 & \textbf{3.35} & 59.80 & 3.71 \\
        AccNet & \textbf{60.8} & \underline{3.58} & \textbf{99.5} & \underline{3.78} & \textbf{95.1} & \underline{3.26} & \textbf{68.31} & \textbf{3.83} \\
    \bottomrule
    \end{tabular}
    }
    \label{sota-balance}
\end{table}

\begin{table}[!htbp]
  \centering
  \caption{ {Comparison of models for the \textbf{best} AP on DAD datasets. TTA@80 and TTA@50 means the value of mTTA at recall equals to 80\% and 50\%, separately. \textbf{Bold} and \underline{underlined} values represent the best and second-best performance in each category. Instances where values are not available are marked with a dash (``-'').}}
    \setlength{\tabcolsep}{1mm}
    \resizebox{\linewidth}{!}{
    \begin{tabular}{ccccccccccc}
    \toprule
    Model &Backbone &Publication  & AP(\%)$\uparrow$ & mTTA(s)$\uparrow$ & TTA@R80(s)$\uparrow$ & TTA@R50(s)$\uparrow$ & AUC$\uparrow$\\
    \midrule 
        ACRA\cite{zeng2017agent} &VGG-16 &{\color{purple}ACCV'16} &51.40 & - & - & - & - \\
        DSA \cite{DSA2016Chan} &VGG-16 &{\color{purple}ACCV'16}  & 63.50 & 1.67 & 1.85 & - & - \\
        UniFormerv2  \cite{li2022uniformerv2} &Transformer &{\color{purple}ICCV'23} &65.24 & - & - & - & - \\
        VideoSwin  \cite{liu2022video} &Transformer &{\color{purple}CVPR'22} &65.45 & - & - & - & - \\
        MVITv2  \cite{fan2021multiscale}&Transformer &{\color{purple}CVPR'21}  &65.65 & - & - & - & - \\
        DSTA \cite{Karim2021stdan} &VGG-16 &{\color{purple}TITS'22}& 66.70 & 1.52 & \textbf{2.39} & \textbf{2.22} & 0.67 \\
        UString \cite{BaoMM2020}&VGG-16 &{\color{purple}ACMMM'20}& 68.40 &1.63 & 2.18 & \underline{1.91} & \underline{0.70} \\
        GSC \cite{wang2023gsc}  &VGG-16 &{\color{purple}TIV'23}& \underline{68.90} & 1.33 & 2.14 & 1.90 & 0.69 \\
        XAI-Accident \cite{karim2022xai} & ResNet50 & {\color{purple}TRR'22}& 64.32 & \textbf{1.80} & 0.68 & 0.64 & 0.65 \\
        \textbf{AccNet} & VGG-16, ZoeDepth &-& \textbf{70.10} & \underline{1.73} & \underline{2.23} & \underline{1.91} & \textbf{0.72} \\
    \bottomrule
    \end{tabular}
    }
    \label{sota-best}
\end{table}

\begin{figure}
    \centering
    \includegraphics[width=0.85\linewidth]{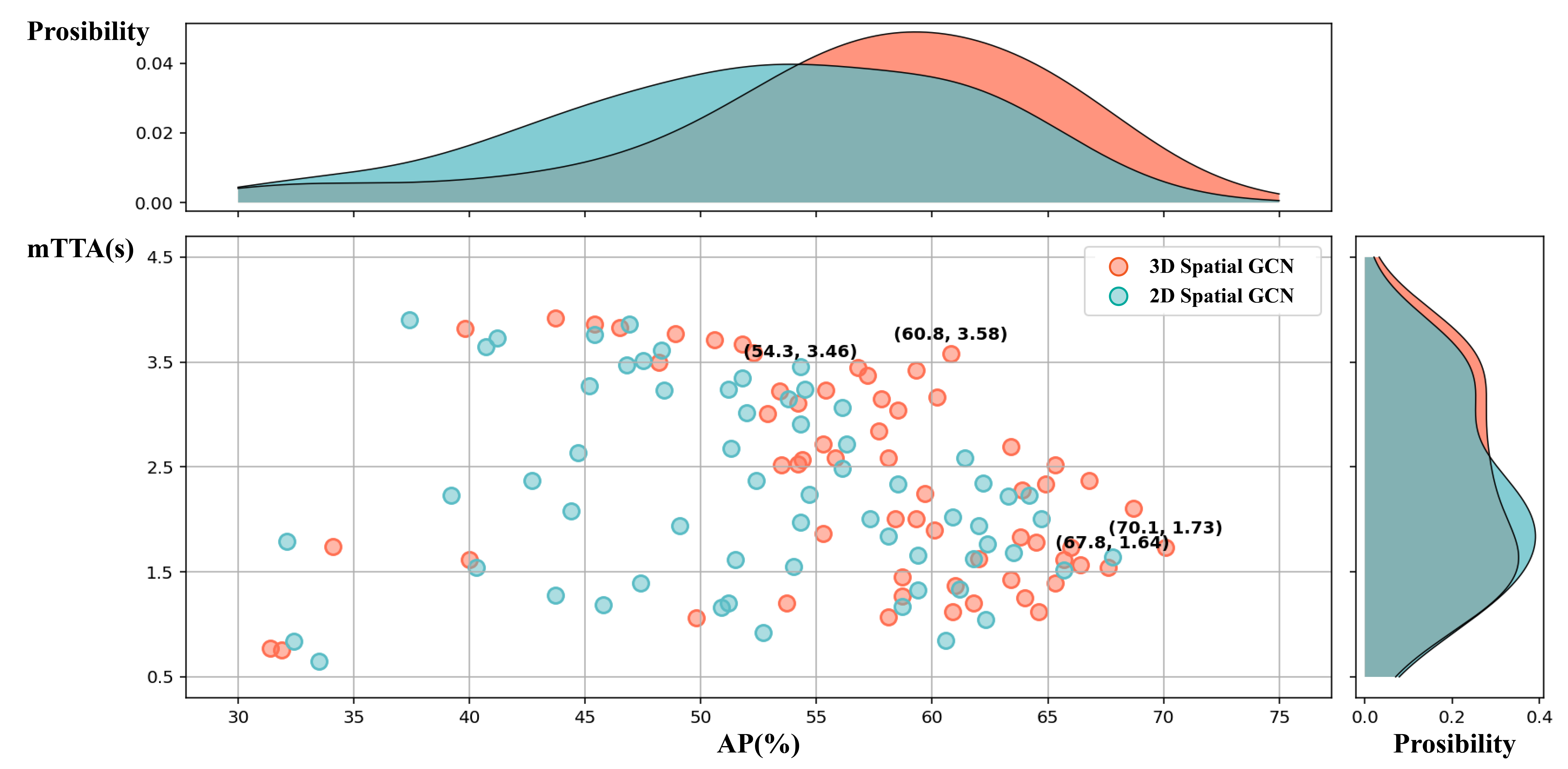}
    \caption{ {Comparison between the 3D Collision and the 2D Collision Modules. Each point in the figure represents the distribution of model performance in mTTA-AP two-dimensional space obtained by training every half epoch using either 2D Spatial GCN or 3D Spatial GCN. At the top and right of the mTTA-AP plot are the probability distributions of mTTA over the AP dimension and AP over the mTTA dimension, respectively. The figure highlights the experimental results showing that the highest AP and a balance between mTTA and AP are achieved with the two types of Spatial GCN.}}
    \label{distribute}
\end{figure}

Our experiments on the DAD, CCD, and A3D datasets demonstrate the exceptional performance of our model on all three datasets, as shown in Table \ref{sota-balance}. These evaluation results indicate an inverse relationship between AP and mTTA, suggesting that a higher AP typically results in a shorter mTTA and vice versa, a trend evident from Fig. \ref{distribute}. The presented results demonstrate a balance between the two metrics, indicating that our model achieved the highest AP values across the datasets, outperforming the second-best by 2.6\% on the DAD dataset. AccNet also surpasses every other model in terms of mTTA. Notably, UString is significantly outperformed by our model in AP by 7.1\% on DAD and 1.9\% on A3D, demonstrating superior overall performance.

 {
To demonstrate the superiority of our model, we compared the best AP among the models. Since there were only small differences between the models on the CCD and A3D datasets, we primarily focused on the DAD dataset. Table 
\ref{sota-best} shows that AccNet achieved the highest AP value, along with the corresponding highest mTTA. Moreover, our model achieved the highest AUC, indicating that it not only effectively distinguishes accident videos but also ensures the highest accuracy. However, it is important to note that our TTA@80 and TTA@50 performance are slightly lower than that of DSTA, indicating that our model takes a more cautious approach in identifying the majority of positive samples. Notably, as comparing the best mTTA without considering AP does not provide meaningful insights, such a comparison is not included in our analysis.}

\subsection{Ablation Studies}
\subsubsection{Ablation Studies of Different Modules}
\begin{table}[t]
  \centering
  \caption{Ablation studies of different modules on DAD datasets. IA, OA, SGM, TA, SM, and AM represent Context Attention, Object Attention, 3D Collision Module, Temporal Attention, Smooth Module, and Accident Module, respectively. TTA@80 means the value of mTTA at recall equals 80\%.}
    \setlength{\tabcolsep}{3mm}
    \resizebox{0.75\linewidth}{!}{
    \begin{tabular}{cccccccccc}
    \toprule
    \multirow{2}[2]{*}{Model} & \multicolumn{6}{c}{Component} & \multicolumn{3}{c}{Metric} \\
    \cmidrule(l{3pt}r{3pt}){2-7} \cmidrule(l{3pt}r{3pt}){8-10} \multicolumn{1}{c}{} & IA & OA & 3D-CM & TA & SM & AM & AP(\%)$\uparrow$ & mTTA(s)$\uparrow$ & TTA@80$\uparrow$ \\
    \midrule 
        A & \ding{56} & \ding{52} & \ding{52} & \ding{52} & \ding{52} & \ding{52} & 56.9 & 3.52 & 3.25 \\
        B & \ding{52} & \ding{56} & \ding{52} & \ding{52} & \ding{52} & \ding{52} & 57.2 & 3.54 & 3.23 \\
        C & \ding{52} & \ding{52} & \ding{56} & \ding{52} & \ding{52} & \ding{52} & 54.3 & 3.46 & 3.26 \\
        D & \ding{52} & \ding{52} & \ding{52} & \ding{56} & \ding{52} & \ding{52} & 58.6 & 3.41 & 3.20 \\
        E & \ding{52} & \ding{52} & \ding{52} & \ding{52} & \ding{56} & \ding{52} & 56.7 & 3.44 & 3.14 \\
        F & \ding{52} & \ding{52} & \ding{52} & \ding{52} & \ding{52} & \ding{56} & 58.4 & 3.32 & 3.19 \\
        original & \ding{52} & \ding{52} & \ding{52} & \ding{52} & \ding{52} & \ding{52} & 60.8 & 3.58 & 3.23 \\
    \bottomrule
    \end{tabular}
    }
    \label{ablantion}
\end{table}

To further validate the contribution of each component within our model, we conducted module ablation studies on the DAD dataset, the results of which are presented in Table \ref{ablantion}. The results of these experiments underscore that the absence of any module results in a decrease in AP, thus confirming the individual contribution of each module to the overall performance of the model. The analysis of Models A and B shows that neglecting the allocation of attention to image and object features leads to a slight improvement in mTTA of +1.7\%, but brings a significant decrease in precision of -6.4\%. Therefore, we decided to retain both modules considering the trade-off. Model C highlights the significant impact of our 3D Collision Module on the experimental results, indicating that traditional methods of constructing graphs using only 2D-pixel coordinates are proven to have significant errors. Additionally, we observed that omitting the 3D Collision module slightly increases TTA@80 by less than 1\%, suggesting that while the module contributes to higher AP, it encourages a more cautious approach to early prediction at higher recall rates. Model E demonstrates the effectiveness of our designed Smooth Module in reducing misclassification caused by potentially dangerous actions in negative videos.

\subsubsection{Ablation Studies of the 3D Collision Module}
To determine if the depth information in the 3D Collision Module (3D-CM) significantly contributes to the model's prediction, we developed a comparative 2D Collision Module (2D-CM). The construction method for the 2D-CM is the same as the 3D-CM, except that it creates a two-dimensional graph representation. Both models were trained and evaluated using the first 30 epochs, with testing performed every half epoch, resulting in 60 test results. The rest of the model architecture and training hyperparameters were kept constant. To make a clearer comparison of the two models, scatter plots of AP vs. mTTA were used, due to the considerable variability during the training phase. 
 {
Figure \ref{distribute} demonstrates that the overall Average Precision (AP) for the 2D-CM is lower than that for the 3D-CM, with a significantly higher variance, indicating increased uncertainty throughout the training process. Experimental results suggest that, when balancing AP and mean Time to Accident (mTTA), the AP and mTTA for the 2D-CM are 54.3\% and 3.46s, respectively, whereas those for the 3D-CM are 60.8\% and 3.58s, marking improvements of 10.7\% and 3.4\%, respectively. Similarly, when comparing the highest AP values, the method using the 3D model also shows enhanced performance, with the highest AP for the 2D-CM at 67.8\%, compared to 70.1\% for the 3D-CM. The 2D-CM struggles to accurately reflect collision position information from the images due to visual parallax issues, leading to challenges in achieving model convergence during training and resulting in greater instability with the 2D-CM approach.}

\subsubsection{Ablation Studies of the Smooth Module}
\begin{table}[tb]
  \centering
  \caption{Ablation studies of Smooth Module. \textbf{Bold} indicates the best experimental results, ``\ding{52}'' indicates that the corresponding conv-deconv is used, and ``-'' indicates that the corresponding conv-deconv is not used.}
    \setlength{\tabcolsep}{3.6mm}
    \resizebox{0.65\linewidth}{!}{
    \begin{tabular}{cccccccc}
    \toprule
    \multirow{2}[2]{*}{Experiment} & \multicolumn{5}{c}{Receptive field (frames)} & \multicolumn{2}{c}{Metric} \\
    \cmidrule(l{3pt}r{3pt}){2-6} \cmidrule(l{3pt}r{3pt}){7-8}  \multicolumn{1}{c}{} & \multicolumn{1}{c}{50} & \multicolumn{1}{c}{20} & \multicolumn{1}{c}{10} & \multicolumn{1}{c}{5} & \multicolumn{1}{c}{2} & AP(\%)$\uparrow$ & mTTA(s)$\uparrow$ \\
    \midrule
        1    & \ding{52} & - & - & - & - & 48.3 & \textbf{4.09} \\
        2    & - & \ding{52} & - & - & - & 50.2 & 3.88 \\
        3    & - & - & \ding{52} & - & - & 52.7 & 3.64 \\
        4    & - & - & - & \ding{52} & - & 53.5 & 3.51 \\
        5    & - & - & - & - & \ding{52} & 51.8 & 3.39 \\
    \midrule
        6     & \ding{52} & \ding{52} & - & - & - & 50.4 & 3.82 \\
        7     & - & \ding{52} & \ding{52} & - & - & 54.0 & 3.71 \\
        8     & - & - & \ding{52} & \ding{52} & - & 57.3 & 3.62 \\
        9     & - & - & - & \ding{52} & \ding{52} & 55.6 & 3.49 \\
    \midrule
        10    & \ding{52} & \ding{52} & \ding{52} & - & - & 53.2 & 3.64 \\
        11    & - & \ding{52} & \ding{52} & \ding{52} & - & \textbf{60.8} & 3.58 \\
        12    & - & - & \ding{52} & \ding{52} & \ding{52} & 58.7 & 3.53 \\
    \midrule
        13    & \ding{52} & \ding{52} & \ding{52} & \ding{52} & - & 54.9 & 3.60 \\
        14    & - & \ding{52} & \ding{52} & \ding{52} & \ding{52} & 56.2 & 3.55 \\
    \bottomrule
    \end{tabular}
    }
    \label{table-smooth}
\end{table}

\begin{figure}[t]
    \centering
    \includegraphics[width=0.85\linewidth]{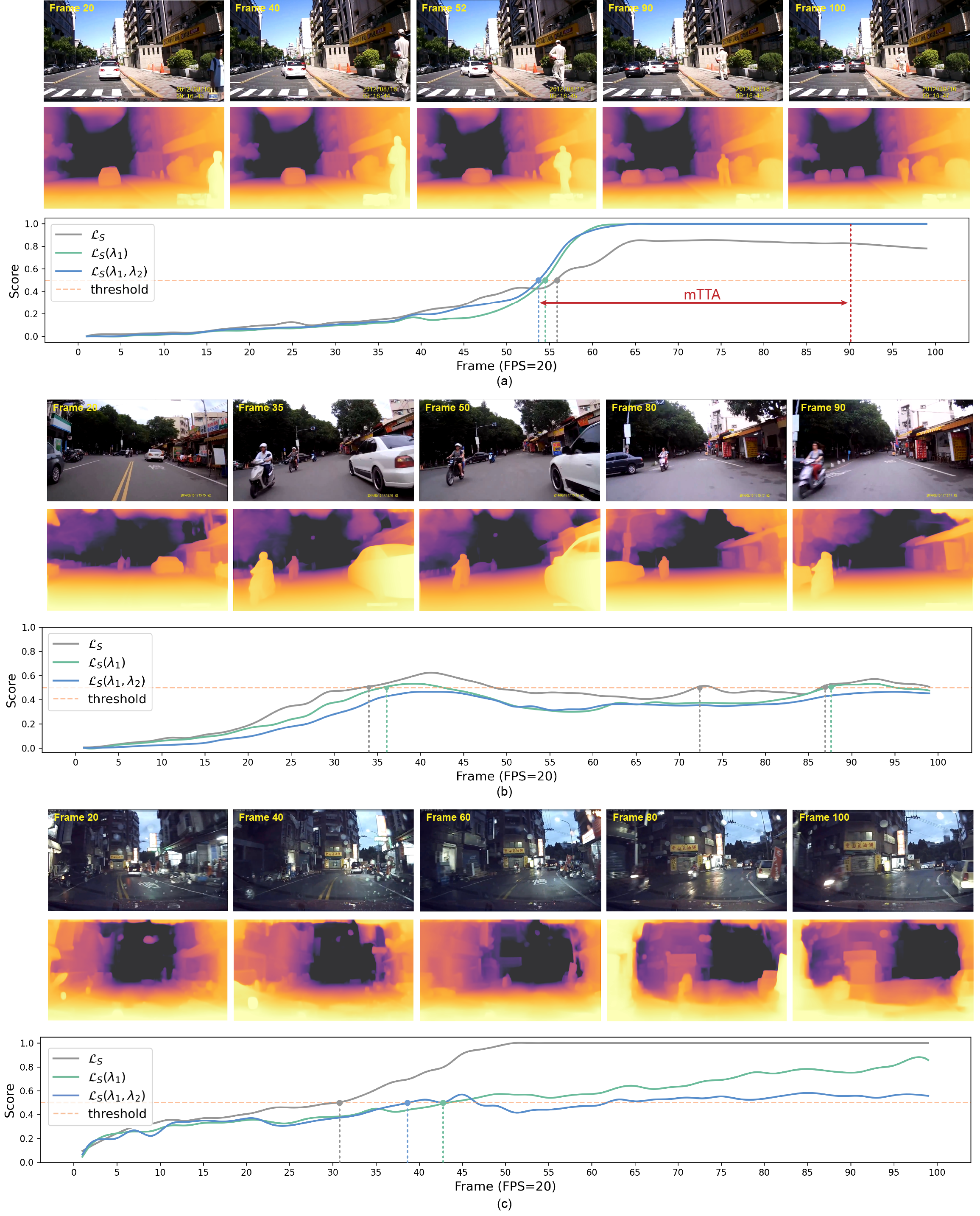}
    \caption{ {Visualization of AccNet's performance in dense urban traffic (a-b) and scenes with low night lighting and rain (c), with the threshold uniformly set at 0.5. Scenes (a-b) are successful accident anticipations, while scene (c) represents a failure case for false positive anticipation.}}
    \label{loss-vis}
\end{figure}

\begin{table}[tb]
  \centering
  \caption{Ablation studies of BE-LEA on two attenuation factors, $f_1$ and $f_2$. \textbf{Bold} indicates the best experimental results, \underline{underline} indicates the second best experimental results within two horizontal lines, and ``-'' indicates that the corresponding coefficient $\lambda_i$ is not used. For example, if $f_1$ is ``-'', then $\lambda_1=1$.}
    \setlength{\tabcolsep}{7.5mm}
    \resizebox{0.65\linewidth}{!}{
    \begin{tabular}{ccccc}
    \toprule
    \multirow{2}[2]{*}{Experiment} & \multicolumn{2}{c}{Coefficient} & \multicolumn{2}{c}{Metric} \\
    \cmidrule(l{3pt}r{3pt}){2-3} \cmidrule(l{3pt}r{3pt}){4-5}  \multicolumn{1}{c}{} & \multicolumn{1}{c}{$f_1$} & \multicolumn{1}{c}{$f_2$} & AP(\%)$\uparrow$ & mTTA(s)$\uparrow$ \\
    \midrule
        1    & - & 10 & 46.9 & \underline{3.53} \\
        2    & - & 50 & 49.2 & 3.45 \\
        3    & - & 100 & 52.1 & 3.37 \\
        4    & - & 150 & \underline{54.4} & 3.29 \\
        5    & - & 200 & 53.6 & 3.26 \\
        \midrule 
        6    & 1 & - & 48.1 & \underline{\textbf{3.74}} \\
        7    & 5 & - & 52.3 & 3.67 \\
        8    & 10 & - & 56.0 & 3.56 \\
        9    & 20 & - & \underline{56.4} & 3.51 \\
        10   & 50 & - & 55.8 & 3.44 \\
        \midrule 
        11   & 10 & 100 & 58.5 & \underline{3.60} \\
        12   & 20 & 100 & \underline{59.7} & 3.57 \\
        13   & 50 & 100 & 57.4 & 3.55 \\
        \midrule 
        14   & 10 & 150 & 59.4 & \underline{3.59} \\
        15   & 20 & 150 & \underline{\textbf{60.8}} & 3.58 \\
        16   & 50 & 150 & 58.8 & 3.53 \\
        \midrule 
        17   & 10 & 200 & 57.2 & \underline{3.54} \\
        18   & 20 & 200 & \underline{58.4} & 3.52 \\
        19   & 50 & 200 & 56.8 & 3.49 \\
    \bottomrule
    \end{tabular}
    }
    \label{loss-co}
\end{table}
\textbf{Multitask adaptive loss function.}
\begin{table}[htbp]
  \centering
  \caption{Ablation studies of BE-LEA on multitask adaptive loss function. \textbf{Bold} indicates the best experimental results within two horizontal lines, \underline{underline} indicates the second best experimental results within two horizontal lines, ``\ding{52}'' indicates that the multitask adaptive method is used, and ``\ding{56}'' indicates that the multitask adaptive method is not used. }
    \setlength{\tabcolsep}{5mm}
    \resizebox{0.7\linewidth}{!}{
    \begin{tabular}{ccccc}
    \toprule
    \multirow{2}[2]{*}{Experiment} & \multirow{2}[2]{*}{Adaptive Method} & \multirow{2}[2]{*}{$\beta$} & \multicolumn{2}{c}{Metric}\\
    \cmidrule(l{3pt}r{3pt}){4-5} & \multicolumn{1}{c}{} & 
    \multicolumn{1}{c}{} & AP(\%)$\uparrow$ & mTTA(s)$\uparrow$ \\
    \midrule
        1    & \ding{56} & $1.0$ & 44.9 & \textbf{3.79} \\
        2    & \ding{56} & $1.0 \times 10^{-1}$ & 48.7 & \underline{3.66} \\
        3    & \ding{56} & $5.0 \times 10^{-2}$ & 50.5 & 3.60 \\
        4    & \ding{56} & $1.0 \times 10^{-2}$ & 51.3 & 3.55 \\
        5    & \ding{56} & $5.0 \times 10^{-3}$ & 54.1 & 3.46 \\
        6    & \ding{56} & $1.0 \times 10^{-3}$ & 56.5 & 3.41 \\
        7    & \ding{56} & $5.0 \times 10^{-4}$ & 57.2 & 3.35 \\
        8    & \ding{56} & $1.0 \times 10^{-4}$ & \underline{58.4} & 3.23 \\
        9    & \ding{56} & $5.0 \times 10^{-5}$ & \textbf{58.6} & 3.17 \\
        10    & \ding{56} & $1.0 \times 10^{-5}$ & 56.4 & 3.02 \\
    \midrule
        Average   & - & - & 53.7 & 3.42 \\
        Variance  & - & - & 21.3 & 0.057 \\
    \midrule
        11    & \ding{52} & $1.0$ & 53.8 & 3.36 \\
        12    & \ding{52} & $1.0 \times 10^{-1}$ & 55.1 & 3.43 \\
        13    & \ding{52} & $5.0 \times 10^{-2}$ & 55.7 & 3.48 \\
        14    & \ding{52} & $1.0 \times 10^{-2}$ & 56.5 & 3.52 \\
        15    & \ding{52} & $5.0 \times 10^{-3}$ & 57.3 & \textbf{3.59} \\
        16    & \ding{52} & $1.0 \times 10^{-3}$ & 59.2 & 3.55 \\
        17    & \ding{52} & $5.0 \times 10^{-4}$ & 60.4 & 3.54 \\
        18    & \ding{52} & $1.0 \times 10^{-4}$ & \textbf{60.8} & \underline{3.58} \\
        19    & \ding{52} & $5.0 \times 10^{-5}$ & \underline{60.6} & 3.49 \\
        20    & \ding{52} & $1.0 \times 10^{-5}$ & 59.9 & 3.41 \\
    \midrule
        Average   & - & - & 57.9 & 3.50 \\
        Variance  & - & - & 6.6 & 0.006 \\
    \bottomrule
    \end{tabular}
    }
    \label{loss-ad}
\end{table}
The Smooth Module in our study consists of three convolution-deconvolution (Conv-Deconv) pairs, each tasked with smoothing operations over different receptive fields. This design allows for the learning of information from different temporal lengths, thereby mitigating the impact of transient dangerous maneuvers on predictions. We conducted tests on different combinations of Conv-Deconv pairs, varying both in number and receptive field size, as detailed in Table \ref{table-smooth}. A comparative analysis of experiments 1-5, 6-9, 10-12, and 13-14 shows that decreasing the receptive field size leads to an increase in AP, albeit at the expense of reduced mTTA values. However, receptive fields that are too small may inversely decrease AP, either because excessive integration of historical data dilutes critical instantaneous information, or because insufficient historical data renders the smoothing module nearly ineffective. This pattern is also consistent with the number of Conv-Deconv pairs used: too few pairs result in uni-dimensional learning of historical data, while too many pairs overgeneralize the result by averaging the outputs of different combinations. As a result, Experiment 11 is chosen for the composition of the Smooth Module, achieving a balanced trade-off between AP and mTTA.

\subsection{Case Studies of the BA-LEA Loss and Qualitative Results}
This study extends the classical loss function optimization from AlaLEA to BA-LEA. The BE-LEA loss depends on two critical components: the $\lambda_1,\lambda_2$ coefficients and the implementation of the multitask adaptive loss function.

 {We train the models separately using $\mathcal{L}_S$, $\mathcal{L}_S(\lambda_1)$, and $\mathcal{L}_S(\lambda_1, \lambda_2)$, where $\mathcal{L}_S$, $\mathcal{L}_S(\lambda_1)$, and $\mathcal{L}_S(\lambda_1, \lambda_2)$ represent the loss function without coefficients, with coefficient $\lambda_1$ only, and with both coefficients $\lambda_1$ and $\lambda_2$, respectively. To make the results clearer, we do not use the multitask adaptive method in this part, since it tends to mitigate the differences in model performance corresponding to different parameter values \( f_1 \) and \( f_2 \). We then visualize the probability values produced by three models. As shown in Fig. \ref{loss-vis} (a), for positive videos, the curve corresponding to $\mathcal{L}_S(\lambda_1)$ is smoother than that of $\mathcal{L}_S$ and closer to 1 within the accident occurrence interval, indirectly leading to earlier accident prediction. Comparatively, the curve of $\mathcal{L}_S(\lambda_1, \lambda_2)$ is closely related to that of $\mathcal{L}_S(\lambda_1)$. In Fig. \ref{loss-vis} (b), for negative videos, both $\mathcal{L}_S(\lambda_1)$ and $\mathcal{L}_S$ curves exceed the threshold, resulting in false positive prediction. This happens because the model predicts a higher probability of an accident around frame 35, where the car rapidly maneuvers between a motorcyclist on the left and a white car on the right. Furthermore, around frame 90, despite the visual evidence suggesting a low probability of an accident, the curves for $\mathcal{L}_S$ and $\mathcal{L}_S(\lambda_1)$ still exceed the threshold. This reflects the ``cumulative'' nature of the model in predicting accident probabilities, where predictions made in later frames tend to give higher probability values, partly due to the inclusion of past frame information via the GRU. The curve for $\mathcal{L}_S(\lambda_1, \lambda_2)$ not only provides a below-threshold judgement around frame 35, but also makes a correct judgement around frame 90 due to its suppressive effect over longer time predictions. In summary, the primary role of the $\lambda_1$ coefficient is to anticipate the timing of accidents in positive videos, while the $\lambda_2$ coefficient functions to suppress false positives in negative videos.}

 {In addition to the successful anticipation cases, we also present a failure scenario, as illustrated in Fig. \ref{loss-vis} (c). This case occurred in a densely populated urban area during a night with heavy precipitation, where visibility was significantly reduced. AccNet erroneously predicted a high likelihood of an accident in this scenario. The error can be attributed primarily to the model's limited ability to accurately capture and interpret the scene's depth information under poor visibility conditions. This challenge is further compounded in dynamically complex environments where rain and darkness obscure critical visual cues. This failure case underscores a crucial aspect of our model's current limitations and highlights an essential avenue for future enhancements. Improving the model's robustness and adaptability in adverse weather conditions will be a focus of our ongoing development. Future iterations of the model will incorporate advanced sensory inputs and machine learning techniques that are better equipped to handle variability in environmental conditions, ensuring more reliable performance across a broader range of scenarios.}

Beyond comparison between different loss functions, we conduct additional ablation experiments to further determine the specific values of the coefficients $\lambda_1, \lambda_2$. The result is shown in Table \ref{loss-co}. Experiments 1-5 show significantly lower performance in both AP and mTTA compared to our SOTA model. In experiments 6-10, omitting $\lambda_2=\frac{t}{f_2}$ results in a slight decrease in AP compared to the SOTA model, but only a minimal decrease in mTTA is observed. These results support our claim that the coefficient $f_1$ primarily reduces false positives and anticipates accident timing, while $f_2$ minimizes false negatives. From experiments 1-10, the optimal settings for $f_1,f_2$ are found to be 20 and 150, respectively, which are then further investigated in subsequent ablation studies.
In experiments 11-13, 14-16 and 17-19 we use identical values of $f_2$. Overall, as $f_2$ increases, AP peaks at $f_2=150$ while mTTA gradually decreases. This trend is consistent with the results of experiments 3-5. Comparing experiments 11, 14, 17 (or experiments 12, 15, 18; 13, 16, 19) under the same $f_1$ value, the highest AP is reached when $f_1=20$, but mTTA decreases as $f_1$ increases. Further observation shows that under the condition that $f_2$ equals 100, 150, 200, the variation of mTTA remains within 0.1s. Considering that the human reaction time is around 0.1s, a marginal improvement of 0.01s or 0.02s in crash prediction does not make a significant contribution; however, a 0.1\% increase in AP could mean 1-2 fewer crash misjudgments, potentially saving lives. Therefore, we choose the coefficients corresponding to the highest AP value, i.e., $f_1=20$ and $f_2=150$.

In this study, we explore the utility of multi-task loss functions, focusing primarily on their ability to autonomously optimize the ratio of loss functions corresponding to different tasks, thus eliminating the need for manual parameter tuning. We conduct experiments to compare scenarios with and without the application of multi-task adaptive conditioning over various loss function ratio coefficients $\beta$. The results of these experiments are presented in Table \ref{loss-ad}. Experiments 1-10 are performed without multi-task adaptive conditioning, while experiments 11-20 are performed with this methodology. It is clearly observed that the application of multi-task adaptiveness resulted in higher means for both AP and mTTA compared to the non-adaptive approach. More importantly, under varying initial coefficients $\beta$, adaptive adjustment lead to a reduced variance in the final results for both AP and mTTA, indicating a more stable generation performance. This suggests that effective results can be achieved without the tedious process of manually adjusting ratio coefficients in ablation studies.

\section{Conclusion}
In this study, we have introduced AccNet, an innovative traffic accident prediction model leveraging Monocular Depth-Enhanced 3D Modeling to precisely extract 3D coordinates of objects from images. This approach has enabled us to construct a novel topological structure for Graph Neural Networks, substantially enhancing the spatial dynamic understanding essential for analyzing dashboard camera footage. The introduction of a new Binary Adaptive Loss for Early Anticipation refines the model's focus on crucial moments, significantly enhancing its predictive accuracy. Furthermore, by employing a multi-task learning strategy, our model adeptly balances the weight between different loss functions, ensuring an optimized gradient descent path and stable training. Evaluated across three benchmark datasets—DAD, CCD, and A3D—AccNet has demonstrated superior performance, outpacing existing models in terms of both Average Precision and mean Time-To-Accident, affirming its exceptional capability in anticipating traffic accidents.

The implications of AccNet's performance are profound, particularly for the development of advanced driver-assistance systems (ADAS) and the future of autonomous driving technologies. By accurately predicting accidents before they occur, AccNet offers the potential to significantly reduce the number of traffic accidents, thereby saving lives and reducing injury severity. Its ability to understand spatial dynamics in real-world scenarios also opens up new avenues for research in visual perception and decision-making processes in autonomous systems.

 {In the future, the integration of AccNet with real-time vehicular systems presents an exciting avenue for enhancing road safety. Future research could explore the adaptation of AccNet to different environmental conditions and its performance in diverse scenarios, including varying weather conditions and urban vs. rural driving environments. Additionally, further refinement of the model to reduce computational requirements without compromising accuracy could facilitate its deployment in real-world applications. Exploring the incorporation of additional data sources, such as lidar and radar, may also enhance the model's robustness and reliability, paving the way for a new era of traffic safety solutions.}

\printcredits

\section*{Acknowledgement}
This research is supported by Science and Technology Development Fund of Macau SAR (File no. 0021/2022/ITP, 0081/2022/A2, 001/2024/SKL), Shenzhen-Hong Kong-Macau Science and Technology Program Category C (SGDX20230821095159012), State Key Lab of Intelligent Transportation System (2024-B001), Jiangsu Provincial Science and Technology Program (BZ2024055), and University of Macau (SRG2023-00037-IOTSC).

% #and 

\bibliographystyle{model1-num-names}
\bibliography{cas-refs}

\end{sloppypar}
\end{document}